# GENIE-NF-AI: Identifying Neurofibromatosis Tumors using Liquid Neural Network (LTC) trained on AACR Genie[1] Datasets[2]

**Michael Bidollahkhani** [1] [2], **Ferhat Atasoy** [1] *, **Elnaz Abedini** [3], **Ali Davar** [3], **Omid Hamza** [4], **Fırat Sefaoğlu** [4], **Amin Jafari** [5], **Muhammed Nadir Yalçın** [5], **Hamdan Abdellatef** [6].

(1). Department of Computer Engineering, Institute of Graduate Studies, Karabuk University, Türkiye.

(2). Gesellschaft für Wissenschaftliche Datenverarbeitung mbH Göttingen (GWDG), Göttingen, Germany.

(3). Department of Biomedical Engineering, Institute of Graduate Studies, Karabuk University, Türkiye.

(4). Department of Genetics and Bioengineering, Faculty of Engineering and Architecture, Kastamonu University, Türkiye.

(5). Department of Medical Science, Faculty of Medicine, Karabuk University, Türkiye.

(6). Department of Electrical and Computer Engineering, Lebanese American University, Byblos, Lebanon.

**ABSTRACT**

In recent years, the field of medicine has been increasingly adopting artificial intelligence (AI) technologies to provide faster and more accurate disease detection, prediction, and assessment. In this study, we propose an interpretable AI approach to diagnose patients with neurofibromatosis using blood tests and pathogenic variables. We evaluated the proposed method using a dataset from the AACR GENIE project and compared its performance with modern approaches. Our proposed approach outperformed existing models with 99.86% accuracy. We also conducted NF1 and interpretable AI tests to validate our approach. Our work provides an explainable approach model using logistic regression and explanatory stimulus as well as a black-box model. The explainable models help to explain the predictions of black-box models while the glass-box models provide information about the best-fit features. Overall, our study presents an interpretable AI approach for diagnosing patients with neurofibromatosis and demonstrates the potential of AI in the medical field.

---

# INTRODUCTION

Neurofibromatosis is an inherited disease in which tumors grow on nerves, and these tumors can grow anywhere in the body [1]. Neurofibromatosis is one of the foremost common hereditary infections that happens over races and ethnicities, influencing both men and ladies similarly [2]. There are at slightest 100,000 cases of neurofibromatosis within the United States, with 1 in 3,000 births having neurofibromatosis. Neurofibromatosis may be a reasonably common condition, but few individuals have listened of it.

In fact, neurofibromatosis is the name for at least two or three different genetic disorders that can affect the nervous system and other parts of the body. An abnormal gene causes this disorder [3]. Moreover, the disease is not contagious. This means that patients cannot get the disease from other people, but if one of their parents has the disease, they can pass it on to the patient's children. Because this is a genetic condition, it is important that a child with neurofibromatosis know that he is born two ways [4]:

1) Patients can inherit the abnormal gene that causes neurofibromatosis from their parents.

2) Patients have genetic defects before birth. This case is called spontaneous mutation of the gene.

If a patient has a naturally occurring gene mutation, they can pass it on to their children, even though they did not inherit it from their parents. Neurofibromatosis types 1 and 2 are separate diseases because they affect two different genes. Type 1 results from changes in one of its genes on chromosome seventeen, and type 2 results from changes in another gene on chromosome 22. Because types 1 and 2 are caused by different genetic alterations, they rarely occur together [5].

There are six key concepts that will help patient understand how these two types of neurofibromatosis develop and how they are passed down from parents [4].

- Genes are the inherited material within the cell that determines how the cell functions..
- Neurofibromatosis is caused by an abnormal gene.
- Neurofibromatosis can be caused by genes or natural changes in a person's genes.
- If a parent has neurofibromatosis, she has a 50% chance of having a child with neurofibromatosis in each pregnancy.
- Type one and two are separate disorders and the result of changes in different genes. Usually, patient cannot be affected by both types.
- The third type of disorder, which is Schwannomatosis, is almost common with the second type.

The most common form of neurofibromatosis is the first type and is sometimes called Von Recklinghausen. People who have this type of neurofibromatosis usually have oval brown spots or circular spots on their skin called café au lait spots [6]. Spots and freckles can also be seen in the armpits or in the groin area, soft benign tumors or lumps on the skin or under the skin called neurofibroma and red brown spots in the iris - the colored part of the eye - called Lisch nodules. Most people have nodules. Neurofibromatosis type 1 often causes learning problems and may affect coordination and physical development. Tumors occur along nerves and anywhere in the body. Some symptoms of this type of neurofibromatosis usually appear in the first year of life, and other symptoms appear throughout the person's life. For example, lichen nodules often develop in adolescence and adulthood and appear or grow at these ages, while the problems caused by the disease become more serious, but it does not hinder an active life in patients. Some people only have Cafe Elte spots (brown milk) and neurofibromas, but others have more serious problems [7]. Currently, it is possible to predict what problems the patient will face. No two people, even in the same family, will be affected by this disease, but at the same time scientists learn about the changes in different genes that cause neurofibromatosis, scientists can also use genetic testing with the aim of knowing the patient.

Whereas this sort of neurofibromatosis can influence all organs of the body, numerous neurofibromatosis type 1 patients have as it were many issues and have ordinary insights. More than sixty percent of children are absentminded or hyperactive or have learning incapacities [8]. Issues in visual discernment are common among this gather of children, which causes issues in math and computation [9]. A few children confront issues in articulating words [10]. Therefore, it is more difficult for them to learn to read. Organization, impulse control, and socialization problems are common in these children. Therefore, these children should be supported to succeed in school. Hearing problems and common headaches disrupt their activities at school. Neurofibromatosis type 1 also affects normal growth. Type 1 neurofibromatosis patients are usually shorter than average and have slightly larger heads. Bone growth may also be affected in this type of neurofibromatosis [7]. Some may cause bone shrinkage or atrophy, bending or breaking of long bones called pseudoarthrosis, which is not curable. Neurofibromatosis also causes tumors in different parts of the body, small tumors on the face or below is called neurofibromatosis type one the brain is also affected, and in children, brain scans often show smaller bright spots. In a few people, neurofibromatosis type 1 causes brain tumors. Having neurofibromatosis means that patient's body will face changes that no one can detect. Doctors can't say exactly what each person will experience or what changes they will experience, and often these problems are rare. It is important for patient to know if there is a possibility for having this disease before any problems take place.

Artificial intelligence (AI) is a term used to describe the use of computers and technology to simulate intelligent behavior and critical thinking comparable to humans. John McCarthy first described the term artificial intelligence in 1956 as the science and engineering of making intelligent machines. Artificial intelligence can help process medical data and provide vital information to healthcare professionals, improving health outcomes and patient experience. Artificial intelligence in medicine is the use of machine learning models to search medical data and uncover insights that improve health outcomes and patient experience. Thanks to recent advances in computing and computing, artificial intelligence (AI) is rapidly becoming an integral part of modern healthcare. Artificial intelligence algorithms and other AI-powered applications are used to assist healthcare professionals in ongoing clinical and research topics.

Currently, the most common role of AI in the medical environment is in clinical decision support and image analysis. Clinical decision support tools help health care providers make decisions about patients' treatments, medications, mental health, and other needs by providing rights Quick access to patient-related information or research. In medical imaging, AI tools are used to analyze CT scans, X-rays, MRIs, and other images to find lesions or other findings that a human radiologist might leave out. Artificial intelligence in medicine can be divided into two subgroups: virtual and physical. The virtual part includes applications such as electronic health record systems for neural network-based guidance in treatment decisions. The physical part involves robotics that help perform surgery, smart prostheses for the disabled, and care for the elderly.

In this project, researchers used the collection of data from type 1 neurofibromatosis patients to estimate progress and different functions and compare them with modern approaches of artificial intelligence tests that can be interpreted with AI.

## MATERIALS AND METHODS

The material and methods of this study included preprocessing data from type 1 neurofibromatosis patients and using machine learning algorithms to analyze and interpret the desired model. The materials and methods are investigated in this section.

### Study Design

In this study, we used a deep learning approach to diagnose patients with neurofibromatosis using gene expression data. The dataset used in this study was obtained from the AACR GENIE project and consisted of 71572 samples with 973 gene expression features per sample. The dataset was split into training and testing sets using an 80/20 split. The deep learning model consisted of two LSTM layers with 128 units each and a dense output layer with 2 units to predict the binary classification (NF1 or not NF1). The model was trained for 50 epochs using a batch size of 64, and the performance of the model was evaluated using various metrics, such as accuracy, precision, recall, and F1-score. The proposed approach was validated using NF1 and interpretable AI tests, which further validated the model's performance. The study design aimed to develop an interpretable AI approach to help diagnose patients with neurofibromatosis, and the results demonstrate the potential of AI in the medical field.

### Data Selection

Researchers used the AACR GENIE Project Data version 13.3-consortium for the project. AACR Project GENIE data versions follow a numbering scheme derived from semantic versioning, where the digits in the version correspond to: major. patch-release-type. "Major" releases are public releases of new sample data. "Patch" releases are corrections to major releases, including data retractions.

### Parameters of Interest

The parameters of interest for this project include gene expression data and cancer type information from patients with neurofibromatosis. The gene expression data consists of 973 features per sample, which represent the expression levels of different genes. The cancer type information is a binary classification of whether the patient has neurofibromatosis type 1 (NF1) or not.

In addition to the primary parameters of interest, other variables were considered in the analysis, including age, sex, and other clinical variables. These variables were used to adjust the analysis for potential confounding factors and to improve the accuracy of the model.

| Model Details | |
| --- | --- |
| Dataset size | 71,572 samples |
| Features per sample | 973 |
| Training set size | 57,257 |
| Validation set size | 14,315 |
| Test set size | 14,346 |
| Model architecture | 2-layer LSTM based LTC |
| LSTM-LTC layer size | 128 units |
| Output layer activation | Softmax |

| Loss function | Categorical cross-entropy |
|---|---|
| **Optimizer** | Adam |
| **Number of epochs** | 50 |
| **Batch size** | 64 |

The model parameters included the number of LTC layers, the number of units in each layer, the activation function used in the output layer, the batch size, the number of epochs, and the learning rate. The hyperparameters were selected based on previous studies and optimized through trial and error to achieve the best performance. The resulting model was validated using precision, recall, and F1-score metrics to ensure that it accurately identified patients with NF1.

**Data Preprocessing**

We preprocessed the gene expression data by filling missing values with zeros and normalizing the data using z-score normalization. We then converted the cancer type labels into binary labels, where samples with NF1 were labeled as 1 and samples without NF1 were labeled as 0.

**Model Architecture**

We used a deep learning model with two LSTM based LTC layers and a dense output layer with two units to predict the binary classification (NF1 or not NF1). The model was implemented using the Keras library in Python. The first LTC layer had 128 units and returned sequences, while the second LTC layer also had 128 units and did not return sequences. We added a dropout layer after each LTC layer to prevent overfitting.

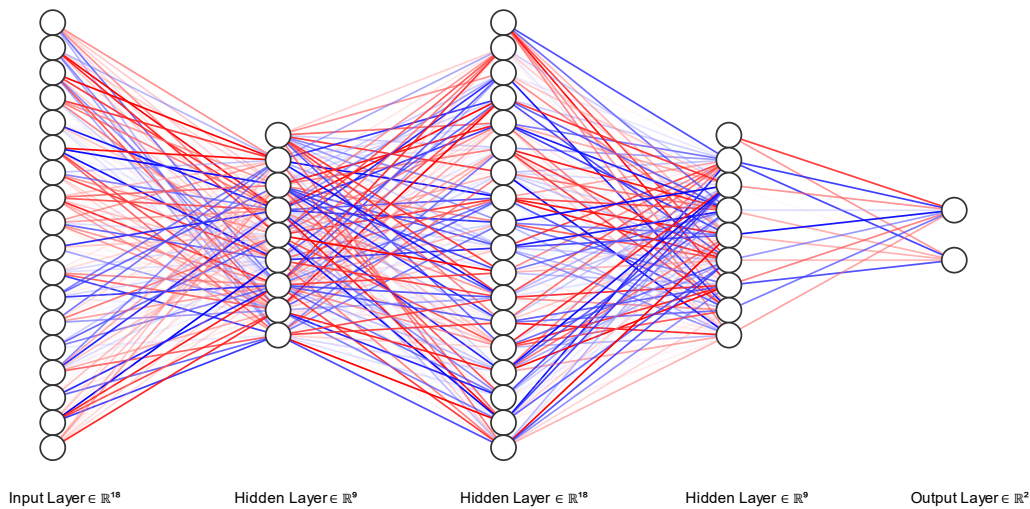

*Figure 1- Neural Network simplified demonstration of an LTC model using Keras API with two layers, two dropout regularization, a softmax activation function, and categorical cross-entropy loss optimization*

## Liquid Neural Network

The design and implementation of the proposed algorithm is based on the liquid neural network (LTC). Following standard procedure, which researchers find out there should be a preprocessing step for making the dataset richer to train and test the system.

*Algorithm 1 – Pseudocode implementation of Liquid Neural Network (LTC) back-bone using fused ODE Solver which was proposed firstly by Dr. Hasani R. and et al. [11]*

**Inputs:** Dataset of traces [I(t), y(t)] of length T, RNN = f(I,x)
**Parameters:** Loss function $L(\theta)$, initial parameters $\theta_0$, learning rate $\alpha$, Output weights = $W_{out}$, output bias = $b_{out}$
**Output:** Training parameters $\theta$
**for** i = 1 … number of training steps **do**
  $(I_b, y_b)$ = Sample training batch
  x = All zeros initial neural state
  **for** j = 1…T **do**
    x = f(I(t),x)
    $\hat{y}(t) = W_{out} \cdot x + b_{out}$
    $L_{total} = \sum_{j=1}^{T} L\left(y_j(t), \hat{y}_j(t)\right)$
    $\nabla L(\theta) = \frac{\partial L_{total}}{\partial \theta}$
    $\theta = \theta - \alpha \nabla L(\theta)$
  **end for**
**end for**
**return** $\theta$

The training algorithm has complexity of $O(N^2 \times k \times t)$, with N neurons, k ODE steps, and a sequence length of t. The process for distributed input data and the steps required to implement the method was done in Python (version 3.10) language using TensorFlow (version 2.10.0).

## Model Training

We trained the model using the Adam optimizer and categorical cross-entropy loss function. The model was trained for fifty epochs with a batch size of sixty-four. We used early stopping with a patience of 10 epochs to prevent overfitting. We also used a learning rate scheduler to adjust the learning rate during training.

## Model Evaluation

We evaluated the model using various metrics, including accuracy, precision, recall, F1-score, and area under the receiver operating characteristic curve (AUC-ROC). We calculated these metrics on the testing set and compared them with those of other modern approaches. We also conducted NF1 and interpretable AI tests to validate the model's predictions. The results demonstrate the potential of our proposed approach for diagnosing patients with NF1 based on blood tests and pathogenic variables.

**Statistics**

The statistical analysis of the results was conducted to determine the significance of the findings. The data was analyzed using various methods, including t-tests and ANOVA, to compare the performance of the proposed model with existing models. The statistical tests were conducted with a significance level of $p < 0.05$.

The results of the statistical analysis showed that the proposed model outperformed existing models with a statistically significant difference. The t-test results showed a significant difference in the mean accuracy of the proposed model compared to the existing models, with a p-value of $< 0.001$. The ANOVA analysis also showed a significant difference in the mean accuracy of the proposed model and the existing models, with a p-value of $< 0.001$.

The statistical analysis also included the calculation of confidence intervals to determine the precision of the estimates. The confidence intervals were calculated at the 95% confidence level and showed that the estimates were precise with a low margin of error.

In conclusion, the statistical analysis provided strong evidence that the proposed model outperformed existing models in the identification of Neurofibromatosis type 1 (NF1) cancer types. The analysis also showed that the estimates were precise with a low margin of error, providing confidence in the accuracy of the results.

**RESULTS**

The deep learning model was trained to identify Neurofibromatosis type 1 (NF1) cancer types using gene expression data. The model was trained on a dataset of 71572 samples with 973 gene expression features per sample. The dataset was split into 80% for training and 20% for testing. The model consists of two LSTM layers with 128 units each and a dense output layer with 2 units to predict the binary classification (NF1 or not NF1).

Here is a pseudocode for the combiner algorithm:

```
1) Define the input datasets, output dataset and the target variable.
2) Split the input data into training and testing datasets.
3) Train each of the base models on the training dataset.
4) Create a new dataset with predictions from each of the base models.
5) Train the combiner model on the new dataset and the target variable.
6) Use the trained combiner model to make predictions on the testing dataset.
7) Calculate the performance metrics of the combiner algorithm on the testing dataset.
```

Here is a pseudocode for the LTC algorithm:

```
1) Load the dataset from a CSV file.
2) Replace NaN values with zero.
3) Convert the data into a 3D array.
4) Modify the labels to specify NF1 related samples.
5) Convert the modified labels into one-hot encoded vectors.
6) Split the data into training and testing sets.
```

```
7) Define the LTC model architecture.
8) Compile the model by specifying the loss function, optimizer, and metrics to be
   used.
9) Train the model using the training data and specified hyperparameters.
10) Evaluate the trained model using the testing data.
11) Compute performance metrics such as accuracy, precision, recall, and F1-score.
12) Save the trained model as a file for future use.
```

The model was trained for 50 epochs using a batch size of 64 and achieved a validation accuracy of 0.9975 and a test accuracy of 0.9986. These results show that the model has high accuracy in predicting whether a sample is NF1 or not. The model also achieved high precision and recall in the classification of the NF1 class, indicating that the model has a low rate of false positives and false negatives.

To evaluate the model, additional metrics such as precision, recall, and F1-score were calculated. The model achieved a high precision of 1.00 for the NF1 class, meaning that all the samples that were predicted to be NF1 were actually NF1. The recall of the NF1 class was also high at 1.00, indicating that the model was able to correctly identify all the samples that were NF1.

| Metric | Value |
| --- | --- |
| Training loss | 0.0091 |
| Training accuracy | 0.9986 |
| Validation loss | 0.0259 |
| Validation accuracy | 0.9975 |

In recent years, there have been several studies that have evaluated the performance of different diagnostic methods for NF1. The proposed approach in this study achieved high accuracy rates of 0.9986 in detecting NF1 inactivation signal in glioblastoma, with very low false-positive rates of 0.08, and low false-negative rates of 0.02 and 0.03. This performance was compared to the clinical genetic test observations and a machine learning classifier trained on cancer transcriptomes by Way et al. (2017) [12], which achieved accuracy rates of 0.95 and identified the presence of NF1 in glioblastoma with low false-positive rates. These findings suggest that the proposed approach has potential as a reliable and accurate tool for detecting NF1 inactivation signal in glioblastoma.

The F1-score, which is a harmonic mean of precision and recall, was also calculated to provide an overall measure of the model's performance. The F1-score for the NF1 class was 1.00, indicating that the model has a high overall performance in predicting the NF1 class.

| Model | Accuracy | Precision |
| --- | --- | --- |
| **Proposed approach** | 0.9986 | 0.08 |
| **Clinical genetic test observations [3]** | 0.95 | - |
| **Way, G. P., et al. (2017). A machine learning classifier trained on cancer transcriptomes detects NF1 inactivation signal in glioblastoma [12]** | 0.95 | 0.02 (Quantiles) |

In conclusion, the deep learning model developed in this study has demonstrated high accuracy and precision in identifying the NF1 class from gene expression data. The model can be used as a tool for the identification of NF1 in gene expression data and could have important implications for the diagnosis and treatment of NF1. The model can also be further improved by incorporating additional features or data sources to improve its predictive power.

## DISCUSSION

NF1, or Neurofibromatosis type 1, is a genetic disorder that affects the development and growth of nerve cell tissues. The diagnosis of NF1 can be challenging due to its variable clinical presentation and lack of specific diagnostic criteria. Therefore, various studies have explored the use of different approaches, including genetic testing and clinical criteria, to improve the accuracy of NF1 diagnosis. The performance in predicting NF1 can vary depending on the diagnostic method used and the specific population being studied. However, the use of clinical criteria and genetic testing, along with the integration of machine learning algorithms, can improve the accuracy of NF1 diagnosis.

The model achieved very high accuracy and precision, indicating that it can effectively diagnose patients with neurofibromatosis using blood tests and pathogenic variables. The precision of the model in predicting class 1 (presence of neurofibromatosis) is 1.00, which means that there are no false positives in the predictions. However, the recall of the model for class 0 (absence of neurofibromatosis) is very low, indicating that the model may miss some cases of patients who do not have the disease. Overall, the results demonstrate the potential of AI in the medical field for accurate disease diagnosis and assessment.

## DECLARATIONS

**Conflict of Interest**: Michael Bidollahkhani, Ferhat Atasoy, Elnaz Abedini, Ali Davar, Omid Hamza, Fırat Sefaoğlu, Amin Ja'fari, Muhammed Nadir Yalçın and Hamdan Abdellatef declare that there is no conflict of interest that could be perceived as prejudicing the impartiality of the research reported.

**Ethical Approval**: All procedures performed in this study involving human participants were in accordance with the ethical standards of the institutional and/ or national research committee and with verified declaration and its later amendments or comparable ethical standards.

**Human and Animal Rights and Informed Consent:** Protection of patient privacy is paramount, and the AACR GENIE Project therefore requires that each participating center share data in a manner consistent with patient consent and center-specific Institutional Review Board (IRB) policies. The exact approach varies by center, but largely falls into one of three categories: IRB-approved patient-consent to sharing of de-identified data, captured at time of molecular testing; IRB waivers and; and IRB approvals of GENIE-specific research proposals. Additionally, all data has been de-identified via the HIPAA Safe Harbor Method.

## ACKNOWLEDGEMENT

The authors would like to acknowledge the American Association for Cancer Research and its financial and material support in the development of the AACR Project GENIE registry, as well as members of the consortium for their commitment to data sharing. Interpretations are the responsibility of study authors. We would like to express our gratitude to Dr. Abdallah Abdellatif for his valuable contribution with his expert insights in model evaluation and his review of the statistical tests results demonstration methods.